\documentclass[letterpaper]{article} 
\usepackage{aaai2027}  
\usepackage[hyphens]{url}  
\usepackage{graphicx} 
\usepackage{natbib}  
\usepackage{caption} 
\usepackage{tcolorbox} 
\tcbuselibrary{breakable,skins} 
\usepackage{algorithm}
\usepackage{algorithmic}
\usepackage{array}
\usepackage{booktabs}
 \usepackage{amsmath}
\usepackage{tabularx}
\usepackage{enumitem}
\newcommand{\tool}{CAPA}
\newcommand{\toole}{CAPA~}

\title{Fewer Clarifications, Better Code: Benchmarking Cross-Session Personalized Ambiguity Adaptation in Coding Assistants}

\author {
    Zijian Xu\textsuperscript{\rm 1, \rm 3}\equalcontrib,
    Wenshuo Zhang\textsuperscript{\rm 2}\equalcontrib,
    Zisen Qin\textsuperscript{\rm 1},
    Rui Sheng\textsuperscript{\rm 2}, 
    Yushi Sun\textsuperscript{\rm 2},
    Huamin Qu\textsuperscript{\rm 2}, 
    Chuhan Shi\textsuperscript{\rm 1}\corresponding
}
\affiliations {
    \textsuperscript{\rm 1}Southeast University\\
    \textsuperscript{\rm 2}The Hong Kong University of Science and Technology\\
    \textsuperscript{\rm 3}The Hong Kong University of Science and Technology (Guangzhou)\\
    zxu477@connect.hkust-gz.edu.cn, wzhangeb@connect.ust.hk, 213230934@seu.edu.cn,
    ysunbp@connect.ust.hk
}

\begin{document}

\maketitle

\begin{abstract}
    AI-assisted coding increasingly translates informal user intent into executable software, yet coding requests often contain ambiguities that recur in user-specific ways across tasks and sessions.
Existing disambiguation methods typically address each ambiguous request in isolation within the current coding session, often through eliciting additional clarification. However, whether resolved session history from the same user can serve as memory for resolving recurring personalized ambiguity in a newly opened session remains underexplored.
We formulate \emph{personalized ambiguity adaptation} as a new task: given a user's previously resolved coding sessions and a new ambiguous request, an assistant should identify the recurring ambiguity pattern, produce the intended executable solution, and minimize clarification. To benchmark this task, we introduce \tool, which characterizes personalized coding ambiguity through six mechanisms and injects these mechanisms into unambiguous executable tasks using a controlled three-stage generation pipeline. \toole contains 600 coding sessions across 60 balanced user--ambiguity cells, including 300 held-out evaluation sessions. We evaluate 12 recent LLMs under no-history and same-user-history conditions using executable success, first-turn success, and turns-to-completion. 
We further study how task difficulty, user identity, and memory management affect adaptation and propose a lightweight user history gating method.
\toole provides a foundation for developing long-term coding assistants that better align generated code with user intent while reducing repeated clarification.
\end{abstract}


\section{Introduction}
AI-assisted coding has become a common way to turn informal intent into executable software. Users ask coding assistants to implement functions, debug failures, adapt existing programs, and modify projects through natural-language dialogue. Yet these requests often remain ambiguous because users may omit details or express intent through recurring, user-specific patterns.
For example, when a user asks to ``normalize'' a feature, the request may refer either to min--max scaling or to z-score standardization. A collaborator familiar with the user may know from prior coding sessions that the user consistently intends the latter one, whereas a coding assistant may select min--max scaling and produce a technically valid but unintended implementation.
A long-term coding assistant should similarly infer such recurring interpretations from prior coding sessions; otherwise, it must repeatedly request clarification or continue making the same incorrect assumption, increasing interaction and user cognitive load.

Recent work addresses important parts of this problem, but largely in isolation. 
Ambiguity research studies whether language models can detect underspecified requests, rewrite them, or ask effective clarification questions~\citep{tanjim2025disambiguation,zhang2024clamber}. These methods generally treat ambiguity as local to the current coding session, while explicit clarification seeks to resolve uncertainty through additional user turns.
Personalization and long-term memory research instead studies how models retrieve, update, and apply user-specific information across sessions~\citep{salemi2024lamp,maharana2024locomo}. However, ambiguity can be user-specific: the same user tends to leave the same kind of information underspecified across different tasks. Whether assistants can use same-user history or personalized memory to resolve such ambiguity without repeated clarification remains underexplored, particularly in open-ended executable coding.
This capability is essential for building long-term assistants that become more accurate and interaction-efficient over time.

To address this gap, we formulate \emph{personalized ambiguity adaptation} as a new task for coding sessions. Given a user's previously resolved sessions and a new ambiguous request that opens a held-out session, an assistant uses the recurring ambiguity pattern revealed by that history to infer the intended implementation. It should ask to clarify only when the history provides insufficient evidence, with the goal of producing a correct executable solution in as few turns as possible.
Unlike conventional ambiguity resolution, which treats each request within its local context, this task evaluates cross-session adaptation to user-specific ambiguity. It motivates assistants that align generated code with user intent while minimizing interruptions from repeated clarification.

To benchmark this task, we introduce \toole(\textbf{C}ross-Session \textbf{A}daptation to \textbf{P}ersonalized \textbf{A}mbiguity), constructed in two steps. First, we adapt the linguistic ambiguity types of \citet{li2024taxonomy} into six mechanisms of personalized coding ambiguity. Each mechanism captures a recurring user-related cause for leaving required implementation information omitted, obscured, or underspecified; these mechanisms are grounded in real coding conversations from WildChat~\citep{zhao2024wildchat}. Second, we instantiate the mechanisms across user profiles and apply a three-stage pipeline to unambiguous HumanEval tasks~\citep{chen2021codex}. The pipeline constructs an ambiguous initial request by controlling task-critical information, expands it into a resolved multi-turn coding session, and validates that the personalized ambiguity pattern remains consistent across sessions. The final benchmark contains 600 coding sessions arranged into 60 balanced user--ambiguity cells, with five resolved history sessions and five held-out evaluation sessions per cell.

We benchmark 12 recent LLMs spanning closed-source frontier, open-access frontier, and smaller open-source models on \toole, and systematically evaluate their personalized ambiguity adaptation using Executable Success (ES), First-Turn Executable Success (FT-ES), and Turns-to-Completion (TTC). We further analyze key aspects of this task, including how adaptation varies with task difficulty, whether history gains depend on correctly matched user identity, and how memory-based history management affects history use. Based on these analyses, we propose same-user history gating as a lightweight method for improving performance.

In summary, our contributions are:
\begin{itemize}
    \item \textbf{A new task for long-term coding assistants.} We formulate personalized ambiguity adaptation as the problem of inferring a user's recurring ambiguity-resolution pattern from resolved coding sessions and transferring it to new executable tasks with minimal clarification.
    \item \textbf{A data generation pipeline.} We introduce \tool, including a six-mechanism coding-oriented taxonomy, a three-stage generation and consistency-validation pipeline, and 600 coding sessions organized into balanced same-user histories and held-out evaluations.
    \item \textbf{An interaction-aware evaluation.} We evaluate 12 models using ES, FT-ES, and TTC, together with controlled studies of task difficulty, user identity, history gating, and memory-based adaptation.
\end{itemize}

\section{Related Work}

\textbf{Ambiguity and clarification.}
Ambiguity occurs when a request has multiple plausible interpretations or lacks information needed to choose among them~\citep{tanjim2025disambiguation}. It may arise from syntax, lexical meaning, or missing conversational context, motivating ambiguity taxonomies and ambiguous question-answering settings~\citep{church1982coping,navigli2009word,schlangen2004causes,liu2023we,min2020ambigqa}. Prior work addresses both ambiguity detection and resolution. Detection methods determine whether to answer or seek information using feature-based classifiers, neural models, or LLM prompting~\citep{trienes2019identifying,dhole2020resolving,guo2021abg,lee2023asking,tanjim2025detecting,kuhn2022clam}; CLAMBER further tests whether LLMs recognize ambiguous needs and ask useful questions~\citep{zhang2024clamber}. Resolution methods rewrite underspecified queries using dialogue or retrieved context~\citep{elgohary2019can,anantha2021open,ma2023query}, cover multiple interpretations in a long-form response~\citep{stelmakh2022asqa,kim2023tree,in2025diversify}, or ask clarifying questions~\citep{aliannejadi2019asking,xu2019asking,zhang2025clarify,kim2024aligning,zhang2025modeling}. In coding, Orchid studies ambiguous requirements in function-level generation~\citep{yang2026assessing}, while ClarifyCodeBench evaluates pre-generation clarification~\citep{fang2026clarifycodebench}. These approaches reduce uncertainty but require additional inference, output, or interaction~\citep{tanjim2025disambiguation}, and resolve requests within the current session or through newly elicited information. \toole instead asks whether an assistant can infer a recurring resolution pattern from a user's prior sessions and reuse it in a new coding session.

\textbf{Personalization and long-term memory.}
Personalized assistants preserve evidence across sessions, infer stable user characteristics, and selectively apply them later. LaMP evaluates history-based personalized classification and generation, while PersonaMem studies user profiling and response selection as personal information evolves across sessions~\citep{salemi2024lamp,jiang2025personamem}. LoCoMo, LongMemEval, and Momento test multi-session recall, temporal reasoning, knowledge updates, and memory-grounded action~\citep{maharana2024locomo,wu2024longmemeval,merin2026momento}. Memory systems use retrieved records, natural-language profiles, or learned representations, increasingly separating episodic evidence from consolidated user patterns~\citep{hu2025memory,cao2026beyond,das2026latent,zhang2026personaagent,hou2026personatree}; practical systems also extract, organize, retrieve, update, and consolidate user information over time~\citep{chhikara2025mem0,xu2026mem,in2026personalize,uddin2026memora,jiang2026learning}. Closest to our setting, APeB infers latent shopping intent from behavioral traces, while PRefine transfers recurring preferences to missing tool-call arguments~\citep{yang2026apeb,yoon2026latent}. Both connect memory and disambiguation but target product choice or schema-bounded tools. \toole instead treats the reusable signal as a personalized ambiguity-resolution pattern and tests its transfer across distinct coding sessions.

\textbf{Interactive coding benchmarks.}
Executable benchmarks objectively assess program correctness. HumanEval and MBPP evaluate function-level generation from explicit specifications, while EvalPlus strengthens test coverage~\citep{chen2021codex,austin2021program,liu2023evalplus}. Interactive benchmarks include SWE-bench for repository-level issue resolution and ConvCodeWorld for conversational generation in reproducible feedback environments~\citep{jimenez2024swebench,han2025convcodeworld}. HumanEvalComm and Orchid specifically examine incomplete, inconsistent, or ambiguous requirements~\citep{wu2025humanevalcomm, yang2026assessing}, but their ambiguity belongs to one task or dialogue rather than a stable cross-session user pattern. \toole combines executable coding, multi-turn sessions, and same-user history. It evaluates both eventual correctness and whether prior sessions reduce clarification.

\section{Benchmark Task}
\label{sec:benchmark-task}

\begin{figure}[t]
    \centering
    \includegraphics[width=\columnwidth]{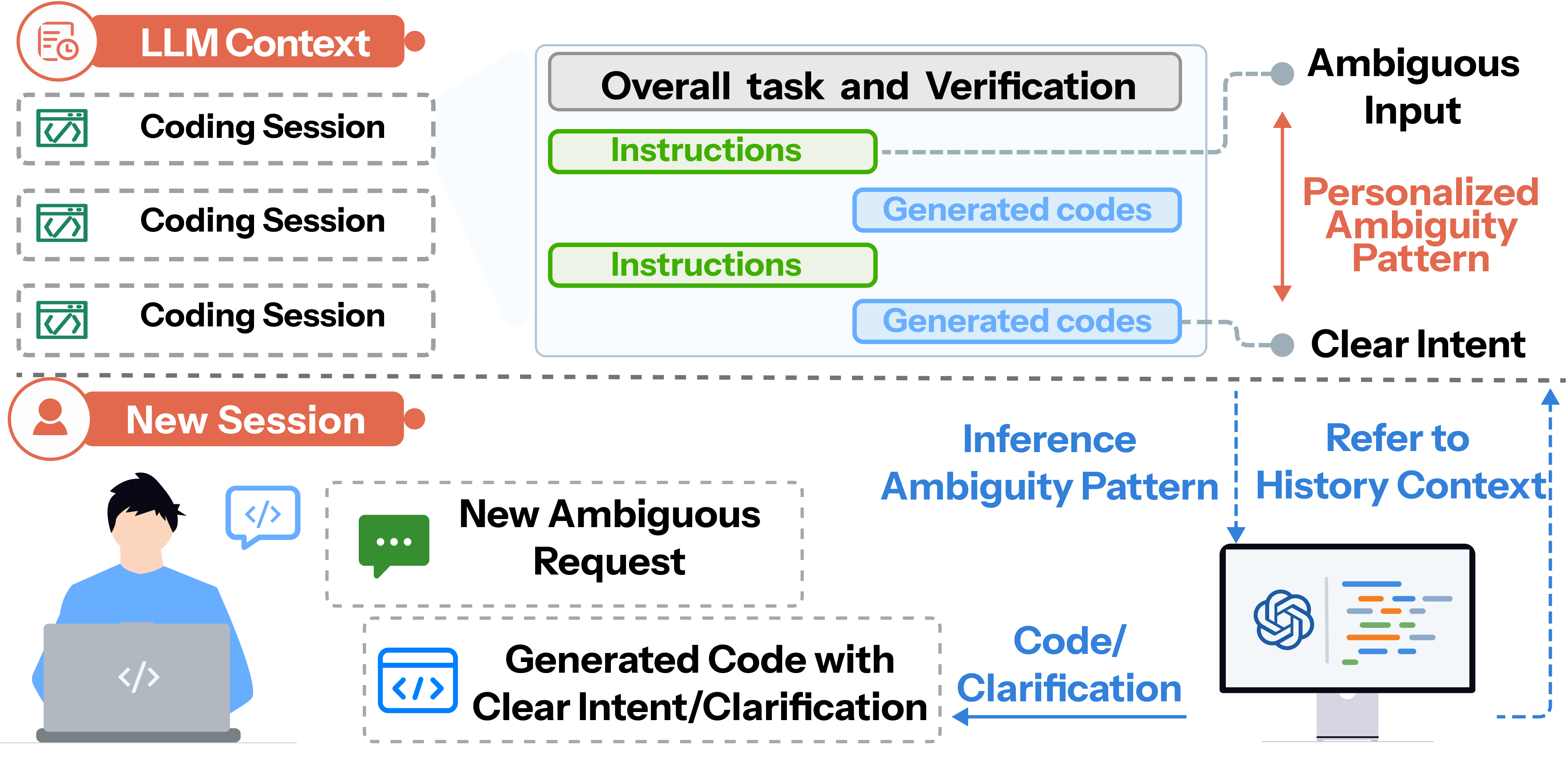}
    \caption{Overview of the history-conditioned personalized ambiguity adaptation task. The assistant receives the public dialogue traces of ($n$) resolved same-user coding sessions and a new ambiguous request that initiates held-out session ($\tau_{i,k}$). It uses cross-session evidence to infer the user's recurring ambiguity pattern, requests clarification only when necessary, and generates code for held-out task $(P_{i,k})$.}
    \label{fig:task}
\end{figure}

\toole evaluates whether a coding assistant can use a recurring user's prior multi-turn coding sessions to resolve personalized ambiguity in a newly opened session. We first define the data hierarchy and notation, and then formulate the history-conditioned evaluation task (Fig.~\ref{fig:task}).

\subsection{Personalized Multi-Session Coding Data}

We distinguish four levels of interaction. An \emph{assistant turn} pairs the latest user message with one assistant response. A \emph{dialogue trace} is the ordered sequence of such turns produced while solving one coding task. A \emph{coding session} is the complete task-centered interaction, including its initial request, dialogue trace, and hidden executable test cases to be evaluated by a hidden judge that determines task completion. A \emph{user trajectory} is an ordered sequence of coding sessions from the same user.

Let user $U_i=(\pi_i,a_i,r_i)$ be characterized by a profile $\pi_i$, a recurring ambiguity mechanism $a_i$, and a personalized resolution pattern $r_i$. The profile controls ambiguity-independent communication characteristics, while $(a_i,r_i)$ specifies how required implementation information is repeatedly obscured and how that ambiguity should be resolved. For the user's $k$-th coding task $P_{i,k}$, the corresponding coding session is
\[
\text{\(\displaystyle \tau_{i,k}=\bigl(D_{i,k},j_{i,k}\bigr)\)},
\]
where $D_{i,k}$ is the public multi-turn dialogue trace produced while solving task $P_{i,k}$, beginning with the ambiguous initial request $u_{i,k}^{(0)}$, and $j_{i,k}$ is the associated set of hidden executable test cases. Writing the trace as
\[
D_{i,k}=\bigl\{(u_{i,k}^{(t-1)},y_{i,k}^{(t)})\bigr\}_{t=1}^{T_{i,k}},
\]
$u_{i,k}^{(t-1)}$ is the latest user message and $y_{i,k}^{(t)}$ is the assistant response at assistant turn $t$; $u_{i,k}^{(0)}$ is the initial request.

The complete trajectory of user $U_i$ is organized as
\[
S_i=\bigl(U_i;
\underbrace{\tau_{i,1},\ldots,\tau_{i,n}}_{\text{resolved history sessions}},
\underbrace{\tau_{i,n+1},\ldots,\tau_{i,n+m}}_{\text{held-out evaluation sessions}}\bigr).
\]
The coding task changes across sessions, whereas $(a_i,r_i)$ remains stable. The first $n$ sessions therefore reveal how the user repeatedly expresses and resolves ambiguity
The following $m$ sessions test whether that personalized pattern transfers to new coding problems.

\subsection{Personalized Ambiguity Adaptation Task}

For a held-out session $\tau_{i,k}$ with $k>n$, the evaluated assistant receives the same-user history
\[
H_i^{(n)}=D_{i,1}\Vert D_{i,2}\Vert\cdots\Vert D_{i,n}
\]
and the new ambiguous request $u_{i,k}^{(0)}$, where $\Vert$ denotes dialogue concatenation. The history is explicitly provided as inference-time context; the task does not assume parameter updates or automatic persistent storage. The assistant must infer the recurring ambiguity pattern evidenced by $H_i^{(n)}$ and apply it within the new multi-turn session.

At assistant turn $t$, let $D_{i,k}^{(<t)}$ denote the dialogue trace accumulated in the current held-out session before that turn. The model produces a response
\[
\text{\(\displaystyle y_{i,k}^{(t)}\)}
=f_{\theta}\!\left(H_i^{(n)},u_{i,k}^{(0)},D_{i,k}^{(<t)}\right),
\]
which either asks a clarification question or contains a candidate code submission. In the former case, the next user message $u_{i,k}^{(t)}$ is added to the dialogue and the interaction continues. In the latter case, the code-extraction rule yields $c_{i,k}^{(t)}$, and the turn succeeds when $J(c_{i,k}^{(t)};j_{i,k})=1$, where the external judge $J$ decides whether the task is completed by combining the results of executing a submission against $j_{i,k}$ with its own assessment of the submitted code against the reference solution. The session terminates at the first successful submission or when it reaches the turn limit.

The desired behavior is to answer directly when same-user history provides sufficient evidence and to clarify only when the intended implementation remains underdetermined. We evaluate this behavior using \textbf{Executable Success} (ES), the proportion of held-out sessions solved within the turn limit; \textbf{First-Turn Executable Success} (FT-ES), the proportion solved at the first assistant turn; and \textbf{Turns-to-Completion} (TTC), the average first-success turn across all held-out sessions, with unsolved sessions assigned the turn-limit value.
Better adaptation improves ES while reducing completion turns, without relying on unsupported guesses.
\section{Data Generation}
\label{sec:dataset-construction}

\begin{figure*}[t]
    \centering
    \includegraphics[width=1\textwidth]{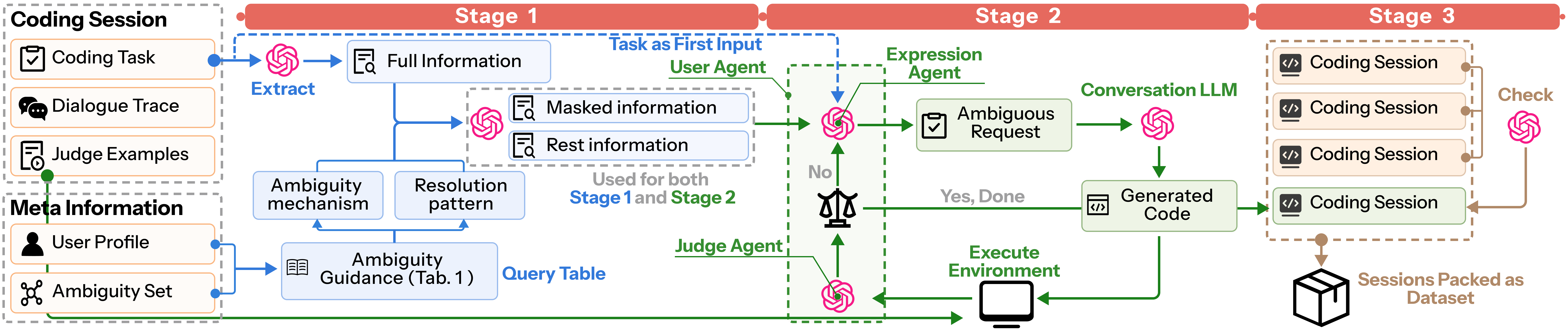}
    \caption{Overview of the \toole data generation pipeline. Stage One combines task requirements with a user profile and an instantiated ambiguity--resolution assignment to create an ambiguous request. Stage Two uses AmbiSimu to produce a coding session through iterative dialogue and execution-based evaluation. Stage Three checks cross-session personalization consistency and assembles accepted sessions into a user trajectory.}
    \label{fig:data-pipeline}
\end{figure*}

This section describes how \toole generates the personalized multi-session data defined in task section.
We first define a coding-oriented ambiguity taxonomy that specifies what kind of uncertainty is repeatedly expressed by a user. Building on this taxonomy, the data generation pipeline proceeds in three stages: constructing the ambiguous initial request for one coding session, generating its multi-turn dialogue trace, and validating personalized consistency across sessions before assembling the user trajectory (Figure~\ref{fig:data-pipeline}).

\subsection{Ambiguity Taxonomy}

To characterize how users repeatedly leave coding intent underspecified, we constructed a coding-oriented taxonomy of personalized ambiguity by adapting the 11 linguistic ambiguity types of \citet{li2024taxonomy}. Rather than classifying ambiguity solely by its surface linguistic form, we reorganized these types around user-related causes of underspecified coding intent. Each category captures a recurring way in which a user understands, assumes, omits, or refers to required implementation information during coding interactions.

Two authors independently mapped the 11 source types to their manifestations in human--LLM coding interactions and grouped the resulting manifestations by their underlying user-related causes. After discussing disagreements, they reached consensus on the category definitions and boundaries, yielding six mechanisms: \emph{domain-cognitive polysemy}, \emph{structural logic misalignment}, \emph{habitual context omission}, \emph{system-boundary misconception}, \emph{conversational context misalignment}, and \emph{implicit constraint under-specification}.
Appendix~\ref{sec:ambiguity-table} provides the full taxonomy. Table~\ref{tab:taxonomy} summarizes each ambiguity mechanism, the type-specific information control used to omit or obscure required implementation information, and the personalized resolution pattern that captures the user-specific interpretation needed to resolve the ambiguity. For each user $U_i$, $a_i$ denotes the assigned ambiguity mechanism, while $r_i$ denotes the corresponding personalized resolution pattern.

\paragraph{Human validation of the taxonomy.}
We conducted a within-subject annotation study with 20 participants recruited through a public call at our university to evaluate the taxonomy. Each participant annotated two sets of 10 items, and adjacent assignments overlapped in a cyclic design so that each of the 200 items received two independent annotations. For every item, participants identified its dominant ambiguity mechanism from our taxonomy and indicated whether it exhibited any additional ambiguity characteristics. 
Inter-annotator agreement yielded a Fleiss' $\kappa$ of 0.66, providing direct empirical evidence that the proposed categories are recognizable in coding requests.


\subsection{Interaction Data Generation Pipeline}

To generate data for the Personalized Ambiguity Adaptation Task, the pipeline operates at two hierarchical levels. At the coding-session level, Stages One and Two transform a clear executable task $P_{i,k}$ into an ambiguous initial request $u_{i,k}^{(0)}$ and its public multi-turn dialogue trace $D_{i,k}$, together forming session $\tau_{i,k}$. At the user-trajectory level, Stage Three checks personalized consistency across sessions before organizing them into $S_i$. This design preserves a stable user-specific ambiguity pattern while allowing the underlying coding tasks and dialogue content to vary.

\textbf{Stage One: Constructing the Ambiguous Initial Request.}
The objective of this stage is to convert a fully specified executable coding task $P_{i,k}$ into the ambiguous initial request $u_{i,k}^{(0)}$ of session $\tau_{i,k}$. Each source task is drawn from HumanEval \citep{chen2021codex} and provides a canonical specification, a function interface, and executable tests.
An initial request is ambiguous when it does not provide all the information required to determine the intended implementation. The six mechanisms above remain the atomic categories of our taxonomy. As a dataset-engineering choice to increase diversity, we assign each user two distinct atomic mechanisms, sampled subject to balanced pairwise coverage, and apply them jointly when rewriting the initial request. Their type-specific information controls and corresponding resolution patterns are combined into one user-level ambiguity--resolution assignment that remains fixed across the user's sessions. In the remainder of the pipeline, $a_i$ and $r_i$ refer to the resulting user-level assignment. Together, its two atomic mechanisms determine which required implementation details are omitted, obscured, or left underspecified.

We denote the information required to determine the intended implementation by $C_{\mathrm{req}}$ and the information retained in the generated initial request by $C_L$. Starting from the canonical task, the generator first extracts the required implementation constraints as $C_{\mathrm{req}}$. It then applies ambiguity mechanism $a_i$ and resolution pattern $r_i$ to remove, obscure, or replace selected constraints, producing $C_L$ such that $C_{\mathrm{req}}\not\subseteq C_L$. Finally, it expresses the remaining information according to profile $\pi_i$, which controls expertise, verbosity, tone, and other ambiguity-independent characteristics, to obtain $u_{i,k}^{(0)}$. The omitted information and its intended interpretation are retained for subsequent dialogue generation and judgment but are not included in the initial request. This procedure creates ambiguity that follows the assigned mechanism and user profile while preserving the original specification and tests $j_{i,k}$ as executable ground truth.

\textbf{Stage Two: Interactive Dialogue Generation.}
The objective of this stage is to expand $u_{i,k}^{(0)}$ into the resolved multi-turn dialogue trace $D_{i,k}$ of one coding session. We construct an interactive environment and model the user agent as two independently inferred components for expression and judgment, ensuring that execution judgment and expression do not effect each other. Unlike conventional dialogue generation, the key requirement is to preserve the assigned personalized ambiguity pattern throughout the session.

To address this requirement, we build the AmbiSimu environment with three components: a \emph{user agent}, a \emph{Conversation LLM}, and an \emph{execution environment}. At each turn, the user agent interacts with the Conversation LLM, which asks a clarification question or generates code $c_{i,k}^{(t)}$. Whenever code is produced, the execution environment compiles and runs it against the session's hidden tests $j_{i,k}$ and returns the result to the user agent. The user agent then continues the dialogue or terminates the coding session; the session also stops when it reaches a predefined turn limit.

Internally, the user agent separates expression from judgment. The \emph{expression agent} receives the task information and generates each user turn by repeatedly applying the context-gap control from Stage One, thereby maintaining the assigned $a_i$, $r_i$, and $\pi_i$ throughout the session. The \emph{judgment agent} invokes the external judge $J$ using two sources of evidence: the code $c_{i,k}^{(t)}$ generated by the Conversation LLM, and the results of compiling and executing that code against $j_{i,k}$. Compilation and execution serve as the primary criterion, while the reference solution and generated code support the assessment of functional agreement. Session $\tau_{i,k}$ terminates successfully when $J(c_{i,k}^{(t)};j_{i,k})=1$ or unsuccessfully when it reaches the turn limit. Only user and Conversation LLM messages are retained in $D_{i,k}$; reference solutions, execution traces, and judgment signals remain hidden. 

\textbf{Stage Three: Cross-Session Consistency Validation.}
The objective of this stage is to keep personalized ambiguity stable across independently generated coding sessions $\tau_{i,k}$ while allowing their underlying tasks and dialogue traces to vary. Stages One and Two first generate multiple task-specific sessions independently for user $U_i$. These sessions use distinct source tasks but share the same profile $\pi_i$, ambiguity mechanism $a_i$, and ambiguity-resolution pattern $r_i$. Stage Three then compares each dialogue trace $D_{i,k}$ with the shared user configuration and the previously accepted trace $D_{i,k-1}$. If the assigned personalized ambiguity is preserved, session $\tau_{i,k}$ is accepted unchanged; only a detected inconsistency triggers revision. The validated sessions $\tau_{i,k}=(D_{i,k},j_{i,k})$ are then assembled into user trajectory $S_i$. The first $n$ sessions form the resolved same-user history, while the following $m$ sessions are reserved for held-out evaluation. This selective validation maintains cross-session consistency without reducing task diversity through unnecessary regeneration.

\paragraph{Dataset statistics.}
The final dataset contains 600 coding sessions arranged into 60 balanced profile--ambiguity-pair cells. Each of the 10 user profiles occurs in six cells, while each of the 15 unordered pairs formed by the six atomic mechanisms occurs in four cells. Each cell contains 10 sessions, yielding 60 sessions per profile. Within every cell, the first five resolved sessions form the history split and the remaining five form the held-out evaluation split. Appendix~\ref{app:dataset_statistics} provides detailed analysis.
Under the no-history reference run, the 300 evaluation sessions comprise 97 simple tasks requiring 1--2 assistant turns (32.3\%), 89 medium tasks requiring 3--4 turns (29.7\%), and 114 complex tasks requiring 5--8 turns (38.0\%).


\section{Experiments}
\label{sec:experiments}

This section provides evaluation protocol and results. 
\subsection{Experimental Setup}


To comprehensively evaluate model performance on \tool, we consider 12 recent LLMs across three groups using model-specific settings, such as temperature, recommended by each model provider. Closed-source frontier models include GPT-5.5, GPT-5.6-Sol, Claude Opus 4.8, Claude Sonnet 4.6, and Gemini 3.5 Flash. Open-access frontier models include DeepSeek V4 Pro, Kimi K2.6, GLM-5.2, and Qwen3.7-Max. We additionally evaluate smaller open-source models, including Llama-3.3-70B-Instruct, Qwen3-8B, and Qwen3.5-27B.

For each model, we follow the coding-session protocol defined in task section on the held-out sessions constructed in dataset section. All settings use the same prompt, eight-turn budget, code-extraction rule, and shared external execution-grounded judge $J$, which evaluates each submission using the hidden test set $j_{i,k}$ from the corresponding session. We report \textbf{Executable Success} (ES), the percentage of held-out sessions solved within the budget; \textbf{First-Turn Executable Success} (FT-ES), the percentage solved directly before clarification; and \textbf{Turns-to-Completion} (TTC), the mean completion turn over all held-out sessions, with an unsuccessful session assigned the maximum value of eight turns.

\subsection{Main Results}

Table~\ref{tab:main-results} reports the main results under no history and same-user history (Five sessions from the same user). Same-user history improves ES for 11 of the 12 models and FT-ES for all 12 models, while reducing average TTC for every model.

\begin{table*}[t]
\centering
\small
\setlength{\tabcolsep}{2pt}
\resizebox{0.85\textwidth}{!}{
\begin{tabular}{llccccccccc}
\toprule
& & \multicolumn{3}{c}{No history} & \multicolumn{3}{c}{Same-user history} & \multicolumn{3}{c}{History gain} \\
\cmidrule(lr){3-5}\cmidrule(lr){6-8}\cmidrule(lr){9-11}
Category & Model & ES $\uparrow$ & FT-ES $\uparrow$ & TTC $\downarrow$
& ES $\uparrow$ & FT-ES $\uparrow$ & TTC $\downarrow$
& $\Delta$ES $\uparrow$ & $\Delta$FT-ES $\uparrow$ & $\Delta$TTC $\uparrow$ \\
\midrule
Closed frontier & GPT-5.5 & 74.3\% & 2.3\% & 4.417 & 84.3\% & 31.0\% & 2.970 & +10.0 pp & +28.7 pp & 1.447 \\
& ChatGPT-5.6-Sol & 79.0\% & 2.7\% & 4.277 & 78.7\% & 18.3\%
 & 3.483 & -0.3 pp & +15.6 pp & 0.793 \\
& Claude Opus 4.8 & 88.0\% & 24.3\% & 2.800 & 90.0\% & 60.3\%
 & 2.113 & +2.0 pp & +36.0 pp & 0.687 \\
& Claude Sonnet 4.6 & 76.0\% & 6.3\% & 3.970 & 83.7\% & 14.0\%
 & 3.380 & +7.7 pp & +7.7 pp & 0.590 \\
& Gemini 3.5 Flash & 79.3\% & 17.0\% & 3.663 & 81.7\% & 38.7\% & 2.927 & +2.3 pp & +21.7 pp & 0.737 \\
\midrule
Open-access & DeepSeek V4 Pro & 76.3\% & 15.3\% & 4.087 & 79.0\% & 17.7\% & 3.667 & +2.7 pp & +2.3 pp & 0.420 \\
& Kimi K2.6 & 77.0\% & 22.3\% & 3.710 & 82.0\% & 32.3\%
 & 3.083 & +5.0 pp & +10.0 pp & 0.627 \\
& GLM-5.2 & 85.3\% & 18.7\% & 3.217 & 89.7\% & 46.7\%
 & 2.457 & +4.4 pp & +28.0 pp & 0.760 \\
& Qwen3.7-Max & 77.7\% & 17.7\% & 3.933 & 82.7\% & 29.7\%
 & 3.120 & +5.0 pp & +12.0 pp & 0.813 \\
\midrule
Compact open-source & Llama-3.3-70B-Instruct & 10.0\% & 5.7\% & 7.450 & 30.0\% & 16.0\% & 6.250 & +20.0 pp & +10.3 pp & 1.200 \\
& Qwen3-8B & 23.3\% & 18.3\% & 6.443 & 27.7\% & 23.0\%
 & 6.133 & +4.3 pp & +4.7 pp & 0.310 \\
& Qwen3.5-27B & 56.0\% & 10.7\% & 5.337 & 74.3\% & 21.3\%
 & 3.990 & +18.3 pp & +10.7 pp & 1.347 \\
\bottomrule
\end{tabular}
}
\caption{Main results under no history and same-user history. ES, FT-ES, and TTC are computed over all sessions; $\Delta$ES and $\Delta$FT-ES denote same-user history minus no history; $\Delta$TTC denotes no-history TTC minus same-user-history TTC.}
\label{tab:main-results}
\end{table*}

\noindent\textbf{\toole remains challenging and far from saturated.} Same-user history benefits nearly all models, yet frontier-model performance spans only 78.7--90.0\% ES and 14.0--60.3\% FT-ES. Claude Opus 4.8, the strongest model, improves from 24.3\% to 60.3\% FT-ES with history, but even this result leaves almost 40\% of sessions requiring clarification or failing after the first response. As an ambiguity-free reference upper bound on the original HumanEval tasks, GPT-5.5 and DeepSeek V4 Pro both achieve 100.0\% ES, 89.0--91.5\% FT-ES, and approximately 1.1 TTC. The large gap from this upper bound indicates considerable room for improving personalized ambiguity resolution.

\noindent\textbf{Open-access frontier models are competitive, while model scale still matters.} With same-user history, closed and open-access frontier models can achieve similar average ES (83.7\% versus 83.4\%) and FT-ES (32.5\% versus 31.6\%). GLM-5.2 is particularly competitive, reaching 89.7\% ES and 46.7\% FT-ES, close to or above most closed models. In contrast, compact models average only 44.0\% ES, although Qwen3.5-27B gains a substantial 18.3 pp from history. Model version alone is also not predictive: GPT-5.5 gains 10.0 pp ES and 28.7 pp FT-ES, whereas ChatGPT-5.6-Sol loses 0.3 pp ES despite gaining 15.6 pp FT-ES.

\noindent\textbf{History improves interaction efficiency more than eventual success.} Averaged across the 12 models, history increases ES by 6.8 pp but FT-ES by 15.6 pp, while reducing TTC by 0.81 turns. GPT-5.5 exemplifies this difference: its ES increases by 10.0 pp, whereas its FT-ES rises by 28.7 pp and TTC decreases by 1.447 turns. Thus, history more clearly helps models identify the intended interpretation and produce executable code earlier than it converts failed sessions into successful ones. 
The different movements show that eventual correctness and interaction efficiency capture complementary aspects of personalized adaptation.

\subsection{Discussion}

\paragraph{Effect of task difficulty.}
As our first research question (\textbf{RQ1}), we investigate how task difficulty affects a model's ability to adapt to personalized ambiguity across held-out sessions.
To answer this question, we use the fixed difficulty partition introduced in dataset section: sessions completed within 1--2 no-history assistant turns are labeled simple, those completed in 3--4 turns are medium, and those taking 5--8 turns are complex, with unsuccessful sessions assigned the eight-turn limit. We evaluate GPT-5.5, DeepSeek V4 Pro, and GLM-5.2 on each subset under same-user history using ES, FT-ES, and TTC. Table~\ref{tab:difficulty-results} summarizes the comparison.

\begin{table}[!htbp]
\centering
\small
\setlength{\tabcolsep}{4pt}
\resizebox{0.8\columnwidth}{!}{%
\begin{tabular}{llccc}
\toprule
Model & Difficulty & ES $\uparrow$ & FT-ES $\uparrow$ & TTC $\downarrow$ \\
\midrule
GPT-5.5 & Simple & 91.75\% & 42.27\% & 2.2165 \\
& Medium & 93.26\% & 31.46\% & 2.5056 \\
& Complex & 71.05\% & 21.05\% & 3.9737 \\
\midrule
DeepSeek V4 Pro & Simple & 88.66\% & 28.87\% & 2.7629 \\
& Medium & 89.89\% & 19.10\% & 3.0899 \\
& Complex & 62.28\% & 7.02\% & 4.8860 \\
\midrule
GLM-5.2 & Simple & 97.94\% & 37.11\% & 2.0619 \\
& Medium & 96.63\% & 41.57\% & 2.0337 \\
& Complex & 75.44\% & 27.19\% & 3.6140 \\
\bottomrule
\end{tabular}%
}
\caption{Performance under same-user history across difficulty levels defined by the no-history coding-session length.}
\label{tab:difficulty-results}
\end{table}

The complex subset remains substantially more challenging even with same-user history: relative to simple sessions, ES drops by 20.7--26.4 pp and TTC increases by 1.55--2.12 turns across the three models, while FT-ES is consistently the lowest. Although performance is not strictly monotonic between simple and medium sessions, all three models perform substantially worse on the complex subset.

\paragraph{Does user identity matter?}
As our second research question (\textbf{RQ2}), we test whether improvements from history reflect genuine user-specific adaptation or merely generic in-context learning from additional sessions. We construct a shuffled-history control by replacing each target user's history with the same number of resolved sessions drawn from other users, while keeping the held-out session and evaluation protocol unchanged. We compare no history, shuffled history, and correctly matched same-user history on GPT-5.5, DeepSeek V4 Pro, and GLM-5.2 using ES, FT-ES, and TTC. Table~\ref{tab:shuffled-results} presents the comparison. Shuffled history may expose generic coding and dialogue patterns, whereas matched history additionally provides personalized ambiguity evidence.

\begin{table}[!htbp]
\centering
\small
\setlength{\tabcolsep}{2.5pt}
\resizebox{0.75\columnwidth}{!}{%
\begin{tabular}{llccc}
\toprule
Condition & Metric & GPT-5.5 & DeepSeek & GLM-5.2 \\
\midrule
No history & ES $\uparrow$ & 74.33\% & 76.33\% & 85.33\% \\
& FT-ES $\uparrow$ & 2.33\% & 15.33\% & 18.67\% \\
& TTC $\downarrow$ & 4.4167 & 4.0867 & 3.2167 \\
\midrule
Shuffled & ES $\uparrow$ & 85.00\% & 77.67\% & 89.00\% \\
& FT-ES $\uparrow$ & 29.00\% & 15.67\% & 34.67\% \\
& TTC $\downarrow$ & 3.0100 & 3.7633 & 2.6433 \\
\midrule
Same-user & ES $\uparrow$ & 84.33\% & 79.00\% & 89.67\% \\
& FT-ES $\uparrow$ & 31.00\% & 17.67\% & 46.67\% \\
& TTC $\downarrow$ & 2.9700 & 3.6667 & 2.4567 \\
\bottomrule
\end{tabular}
}
\caption{Effect of shuffled and correctly matched user history across three models.}
\label{tab:shuffled-results}
\end{table}

Even when user-specific information is removed through shuffling, models still benefit from resolved-session context, with ES improving by up to 10.67 pp over no history. This indicates that generic coding and dialogue patterns can resolve part of the ambiguity. Restoring correctly matched same-user history yields further consistent gains in FT-ES (2.0--12.0 pp) and TTC (0.04--0.19 turns), although its effect on ES is small and mixed. These additional gains show that models can exploit personalized ambiguity-resolution patterns beyond generic contextual learning.

\paragraph{Can existing memory methods strengthen history use?}
As our third research question (\textbf{RQ3}), we investigate whether memory-based history management can improve how models exploit cross-session history. We compare two existing methods, mem0~\citep{chhikara2025mem0} and A-mem~\citep{xu2026mem}, with our \emph{same-user history gating} approach described in the paragraph below. The existing methods retrieve, summarize, or organize evidence from resolved same-user sessions instead of directly inserting the complete raw history. We evaluate each method with GPT-5.5, DeepSeek V4 Pro, and GLM-5.2, reporting ES, FT-ES, and TTC. Table~\ref{tab:memory-results} summarizes the comparison.

\begin{table}[!htbp]
\centering
\small
\setlength{\tabcolsep}{3pt}
\resizebox{0.8\columnwidth}{!}{%
\begin{tabular}{llccc}
\toprule
History method & Model & ES $\uparrow$ & FT-ES $\uparrow$ & TTC $\downarrow$ \\
\midrule
mem0 & GPT-5.5 & 85.67\% & 11.67\% & 3.3133 \\
& DeepSeek V4 Pro & 73.33\% & 9.33\% & 4.3000 \\
& GLM-5.2 & 83.67\% & 18.67\% & 3.5267 \\
\midrule
A-mem & GPT-5.5 & 85.67\% & 29.33\% & 2.9800 \\
& DeepSeek V4 Pro & 74.67\% & 11.67\% & 4.2600 \\
& GLM-5.2 & 81.33\% & 22.67\% & 3.3133 \\
\midrule
Ours & GPT-5.5 & 83.33\% & 44.33\% & 2.7933 \\
& DeepSeek V4 Pro & 79.00\% & 18.33\% & 3.7333 \\
& GLM-5.2 & 90.67\% & 51.33\% & 2.2800 \\
\bottomrule
\end{tabular}
}
\caption{Comparison of memory-based history management.}
\label{tab:memory-results}
\end{table}

Neither general-purpose memory system consistently improves over raw same-user history: both underperform it across all three metrics for DeepSeek V4 Pro and GLM-5.2, while offering only mixed trade-offs for GPT-5.5. One likely reason is an objective mismatch. Mem0~\cite{chhikara2025mem0} focuses on extracting, updating, and retrieving facts, whereas A-mem~\cite{xu2026mem} organizes memories into linked, evolving notes; neither explicitly identifies recurring ambiguity-resolution patterns or requires the model to decide whether the retrieved evidence supports direct implementation. General memory management therefore does not necessarily translate into effective personalized disambiguation, motivating the task-specific gating approach below. Our approach explicitly checks for such ambiguity-resolution evidence and achieves the highest FT-ES and lowest TTC among the three methods for all evaluated models.

\paragraph{A lightweight same-user history gating method.}

\begin{figure}[!htbp]
    \centering
    \includegraphics[width=\columnwidth]{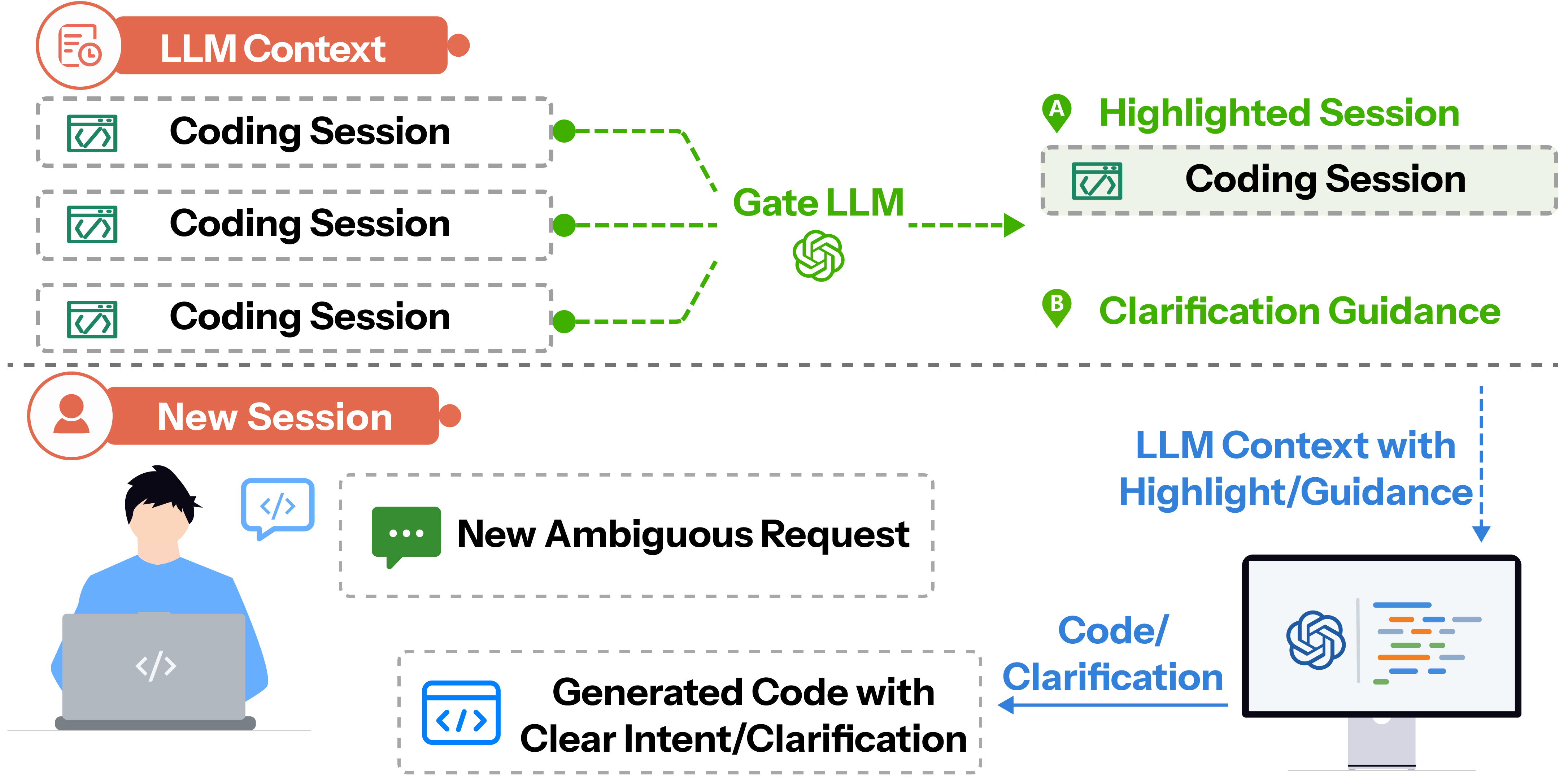}
    \caption{Overview of same-user history gating. 
    }
    \label{fig:gated-history}
\end{figure}

Resolved same-user sessions can reveal recurring personalized ambiguity, yet base models may underuse this evidence and clarify unnecessarily. 
Our parameter-free workflow introduces a gate LLM that reviews the resolved history for consistent ambiguity--resolution evidence. 
When such evidence is sufficient, the gate highlights the most informative prior session; otherwise, it provides clarification guidance indicating what remains unresolved. 
The highlighted evidence or guidance is added to the LLM context together with the new ambiguous request, enabling the base LLM to either generate code directly or request necessary clarification (Figure~\ref{fig:gated-history}).

\textbf{Evaluation.} Compared with raw same-user history, our gating method consistently improves FT-ES by 0.66--13.33 pp while keeping ES within $\pm$1.0 pp across all three models. It also reduces TTC for GPT-5.5 and GLM-5.2, demonstrating that explicitly checking ambiguity-resolution evidence generally enables earlier completion without substantially affecting eventual success.

\begin{table}[t]
\centering
\small
\setlength{\tabcolsep}{2.5pt}
\resizebox{0.75\columnwidth}{!}{%
\begin{tabular}{llccc}
\toprule
Condition & Metric & GPT-5.5 & DeepSeek & GLM-5.2 \\
\midrule
No history & ES $\uparrow$ & 74.33\% & 76.33\% & 85.33\% \\
& FT-ES $\uparrow$ & 2.33\% & 15.33\% & 18.67\% \\
& TTC $\downarrow$ & 4.4167 & 4.0867 & 3.2167 \\
\midrule
Same-user & ES $\uparrow$ & 84.33\% & 79.00\% & 89.67\% \\
& FT-ES $\uparrow$ & 31.00\% & 17.67\% & 46.67\% \\
& TTC $\downarrow$ & 2.9700 & 3.6667 & 2.4567 \\
\midrule
History gated & ES $\uparrow$ & 83.33\% & 79.00\% & 90.67\% \\
& FT-ES $\uparrow$ & 44.33\% & 18.33\% & 51.33\%\\
& TTC $\downarrow$ & 2.7933 & 3.7333 & 2.2800 \\
\bottomrule
\end{tabular}
}
\caption{Comparison of no history, raw same-user history, and same-user history gating.}
\label{tab:gated-results}
\end{table}
\section{Conclusion}

Coding assistants often face recurring user-specific ambiguity yet fail to infer it from prior sessions, resulting in repeated clarification or misaligned implementations. We formulate \emph{personalized ambiguity adaptation} as a new task for long-term coding assistants: inferring a user's recurring ambiguity pattern from resolved sessions and applying it to a new session with minimal clarification. We introduce \tool, a benchmark with six coding-oriented ambiguity mechanisms and 600 executable sessions organized into same-user histories and held-out evaluations. Our evaluation of 12 recent LLMs measures executable correctness and interaction efficiency, showing that cross-session history can improve task completion and reduce clarification. This task and benchmark support coding assistants that better align generated code with user intent while minimizing interruption.

\bibliography{refs}

\clearpage
\appendix

\clearpage
\twocolumn[
\begin{minipage}{\textwidth}
\section{Dataset Balance and Difficulty Analysis}
\label{app:dataset_statistics}

This section summarizes the controlled composition of \toole and reports
reference-run statistics for split comparability and held-out difficulty.

\vspace{0.25em}
\centering
\IfFileExists{figures/capa_dataset_balance_difficulty.png}
  {\includegraphics[width=\textwidth]{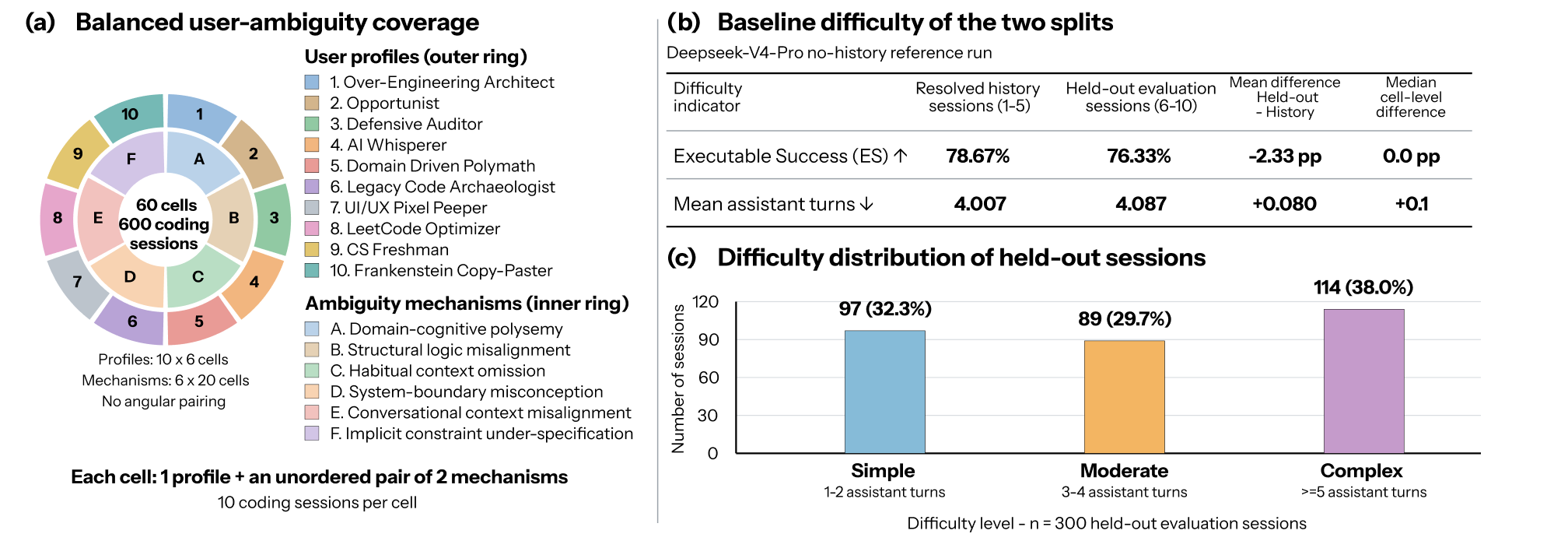}}
  {\includegraphics[width=\textwidth]{capa_dataset_balance_difficulty.png}}
\captionof{figure}{Dataset balance and task difficulty.
  (a) \toole comprises 600 sessions in 60 balanced cells defined by ten user
  profiles and pairs drawn from six ambiguity mechanisms; angular adjacency
  does not encode pairings.
  (b) Resolved history and held-out evaluation sessions have comparable
  no-history baseline difficulty.
  (c) The 300 held-out sessions are partitioned into 97 simple, 89 moderate,
  and 114 complex sessions.}
\label{fig:dataset_balance_difficulty}
\end{minipage}
]

\begingroup
\raggedbottom
\vspace{0.2em}

\noindent\textbf{Balanced user--ambiguity coverage.}
Figure~\ref{fig:dataset_balance_difficulty}(a) shows 600 sessions organized
into 60 cells, with ten sessions per cell. Each of the ten profiles occurs
in six cells. The six mechanisms form 15 unordered pairs, each represented
by four cells, so every mechanism participates in 20 cells. Equal sectors
show aggregate balance; angular position does not encode the pairings used
in individual cells.

\smallskip
\noindent\textbf{Composite instantiation and marginal balance.}
The six mechanisms remain atomic taxonomy categories. For dataset
construction, each cell combines one profile with one mechanism pair, and
both mechanisms are jointly instantiated in every session in that cell.
Their type-specific information controls and corresponding resolution
patterns are combined into a single user-level assignment that remains
fixed across the cell's ten sessions. Consequently, each atomic mechanism
has equal marginal coverage of 20 cells, 200 sessions overall, and 100
held-out sessions. Mechanism-level analyses therefore use multi-label
marginal aggregation: each session contributes to both atomic categories
in its assigned pair.

\smallskip
\noindent\textbf{Design consideration.}
Combining two atomic mechanisms increases construction diversity and
captures requests in which multiple information gaps coexist. However,
performance attributed marginally to one mechanism may also reflect its
paired mechanism and their interaction. The mechanism-level results should
therefore be interpreted as balanced marginal associations rather than
causal effects of isolated mechanisms.

\smallskip
\noindent\textbf{Comparability of the two session splits.}
The first five sessions in each cell form the resolved history and the last
five form the held-out evaluation set. Under the same DeepSeek V4 Pro no-history
reference run, information from the resolved history cannot affect the
comparison. ES is 78.67\% and 76.33\%, while mean assistant turns are 4.007
and 4.087, respectively. The held-out-minus-history differences are $-2.33$
percentage points and $+0.080$ turns; median cell-level differences are 0.0
percentage points and $+0.1$ turns. These small differences indicate that
the held-out sessions are not systematically easier or harder than the
resolved history sessions.

\smallskip
\noindent\textbf{Difficulty distribution of held-out sessions.}
We label sessions requiring one or two assistant turns as \emph{simple},
three or four as \emph{moderate}, and at least five as \emph{complex}.
Figure~\ref{fig:dataset_balance_difficulty}(c) contains 97 simple (32.3\%),
89 moderate (29.7\%), and 114 complex sessions (38.0\%). These no-history
reference labels remain fixed in all subsequent analyses.

\subsection{Blank-Control Experiment}
\label{app:blank_control}

\noindent\textbf{Motivation.}
The blank-control experiment tests whether the underlying programming tasks
are already difficult without the ambiguity introduced by \tool. If models
solve the original tasks reliably without user simulation or information
deletion, degradation in the full benchmark is more plausibly attributable
to ambiguity resolution and multi-turn interaction.

\smallskip
\noindent\textbf{Experimental setting.}
We use all 164 tasks from the official HumanEval benchmark with their
complete original specifications. No ambiguity mechanism, persona
simulation, or deliberate information deletion is applied. The model first
generates a candidate solution from the full task prompt. If it fails, the
debugger returns the concrete execution or test failure and the model revises
its code. This generate--test--feedback--revise process continues until all
tests pass or the model reaches eight submissions.

\begin{center}
\begin{minipage}{\columnwidth}
  \centering
  \footnotesize
  \captionof{table}{Blank-control results on 164 HumanEval tasks. ES is final
  success, FT-ES first-try success, and TTC mean turns; best values are bold.}
  \label{tab:blank_control}
  \resizebox{\columnwidth}{!}{%
    \begin{tabular}{lccc}
      \toprule
      Model & ES $\uparrow$ & FT-ES $\uparrow$ & TTC $\downarrow$ \\
      \midrule
      ChatGPT-5.5
        & \textbf{100.00\%}
        & 89.02\% (146/164)
        & 1.122 \\
      DeepSeek V4 PRO
        & \textbf{100.00\%}
        & \textbf{91.46\% (150/164)}
        & \textbf{1.098} \\
      \bottomrule
    \end{tabular}%
  }
\end{minipage}
\end{center}

\smallskip
\noindent\textbf{Results and interpretation.}
Both models eventually solve all 164 tasks, reaching 100.00\% ES.
DeepSeek V4 PRO obtains the higher FT-ES, solving 150 tasks on the first
attempt compared with 146 for ChatGPT-5.5, and requires slightly fewer turns
on average. These results show that the original tasks are almost fully
solvable under complete specifications and direct debugger feedback.
Therefore, the principal challenge evaluated by \toole is not basic code
generation alone, but recovering and resolving missing, ambiguous, or
misaligned user requirements during interaction.

\endgroup

\twocolumn[{
\begin{minipage}{\textwidth}
\section{Analysis by Personalized Ambiguity Mechanism}
\label{sec:ambiguity-table}

\subsection{Detailed Mechanism Descriptions}
Table~\ref{tab:taxonomy} expands the coding-oriented ambiguity taxonomy introduced in Dataset construction section. For each mechanism, it presents the linguistic ambiguity types from which it is adapted, the information-control operation used to construct ambiguous coding requests, and the personalized resolution pattern required to recover the intended implementation.

\begin{center}
\centering
\scriptsize
\setlength{\tabcolsep}{4pt}
\renewcommand{\arraystretch}{1.08}
\begin{tabular}{
    @{}
    >{\raggedright\arraybackslash}p{0.18\textwidth}
    >{\raggedright\arraybackslash}p{0.15\textwidth}
    >{\raggedright\arraybackslash}p{0.28\textwidth}
    >{\raggedright\arraybackslash}p{0.34\textwidth}
    @{}
}
\toprule
Ambiguity mechanism ($a_i$) & NLP triggers adapted & Type-specific information control & Personalized resolution pattern ($r_i$) \\
\midrule
Domain-cognitive polysemy & Lexical, idiomatic & Replace a standard implementation concept with user- or project-specific terminology. & A stable mapping from the personalized term or shorthand to its intended implementation action. \\
\addlinespace[2pt]
Structural logic misalignment & Syntactic, scopal, collective/distributive & Remove or flatten explicit logical relations, scope, grouping, or thresholds. & The user's recurring interpretation of the omitted logical structure, such as scope or batch-versus-element behavior. \\
\addlinespace[2pt]
Habitual context omission & Elliptical & Omit task context that the user treats as an established convention. & The defaults, input conventions, schemas, or entry points that the user consistently assumes. \\
\addlinespace[2pt]
System-boundary misconception & Presuppositional & Omit the step needed to obtain external state and presuppose that it is already observable. & The user's recurring assumption about the assistant's access to files, logs, runtime state, credentials, or tools. \\
\addlinespace[2pt]
Conversational context misalignment & Coreferential, type/token & Replace an explicit entity, version, example, or decision with a context-dependent reference. & A stable mapping from the reference to the intended item in the user's prior dialogue context. \\
\addlinespace[2pt]
Implicit constraint under-specification & Implicative, generic/non-generic & Retain the main behavior while omitting secondary implementation constraints. & The robustness, validation, safety, or edge-case requirements that the user consistently expects implicitly. \\
\bottomrule
\end{tabular}
\captionof{table}{Coding-oriented ambiguity taxonomy derived from the linguistic ambiguity types of \citet{li2024taxonomy}. For each user, $a_i$ specifies how required information is made ambiguous, while $r_i$ records the personalized interpretation that resolves it.}
\label{tab:taxonomy}
\end{center}
\end{minipage}
}]

\subsection{Effect of Ambiguity Mechanism on Task Performance}
We further examine whether GPT-5.5 performs uniformly across the six mechanisms under same-user history. Because each held-out session contains two jointly applied mechanisms, we use multi-label marginal aggregation: each session contributes to both corresponding categories. The results are reported in Table~\ref{tab:mechanism-performance}.
\begin{center}
\scriptsize
\setlength{\tabcolsep}{2.5pt}
\renewcommand{\arraystretch}{1.08}
\begin{tabular}{
    @{}
    >{\raggedright\arraybackslash}p{0.43\columnwidth}
    rrrr
    @{}
}
\toprule
Ambiguity mechanism & N & ES $\uparrow$ & FT-ES $\uparrow$ & TTC $\downarrow$ \\
\midrule
Domain-cognitive polysemy & 100 & 84.0\% & 24.0\% & 3.020 \\
Structural logic misalignment & 100 & 87.0\% & \textbf{38.0\%} & 2.760 \\
Habitual context omission & 100 & \textbf{88.0\%} & 35.0\% & \textbf{2.630} \\
System-boundary misconception & 100 & 81.0\% & 27.0\% & 3.300 \\
Conversational context misalignment & 100 & 85.0\% & 35.0\% & 2.700 \\
Implicit constraint under-specification & 100 & 81.0\% & 27.0\% & 3.410 \\
\bottomrule
\end{tabular}
\captionof{table}{GPT-5.5 performance across ambiguity mechanisms under same-user history. Because each held-out session contains two mechanisms, it contributes to both corresponding categories; ES, FT-ES, and TTC are marginal statistics over all sessions containing each mechanism.}
\label{tab:mechanism-performance}
\end{center}

Across mechanisms, ES ranges from 81.0\% to 88.0\%, FT-ES ranges from 24.0\% to 38.0\%, and TTC ranges from 2.630 to 3.410 turns. Habitual context omission has the highest ES and lowest TTC, structural logic misalignment has the highest FT-ES, and implicit constraint under-specification has the highest TTC. Overall, performance is broadly comparable across the six mechanisms.
\ifcsname promptbox\endcsname
\else
\newtcolorbox{promptbox}[2][]{
  enhanced,
  breakable,
  colback=gray!7,
  colframe=gray!72!black,
  colbacktitle=gray!72!black,
  coltitle=white,
  fonttitle=\bfseries,
  title={#2},
  title after break={#2\space (continued)},
  boxrule=0.9pt,
  arc=2mm,
  outer arc=2mm,
  left=7mm,
  right=7mm,
  top=3.5mm,
  bottom=3.5mm,
  toptitle=1.5mm,
  bottomtitle=1.5mm,
  lefttitle=7mm,
  righttitle=7mm,
  before skip=7pt,
  after skip=7pt,
  fontupper=\small,
  #1
}
\fi

\newlength{\promptoutertitleskip}
\setlength{\promptoutertitleskip}{2pt}
\newlength{\promptinnertitleskip}
\setlength{\promptinnertitleskip}{4pt}

\newcommand{\promptprimaryheading}[1]{%
  \par\vspace{\promptoutertitleskip}%
  \noindent{\large\bfseries #1\par}%
  \nobreak\vspace{\promptoutertitleskip}%
}

\newcommand{\promptsecondaryheading}[1]{%
  \par\vspace{\promptoutertitleskip}%
  \noindent{\bfseries #1\par}%
  \nobreak\vspace{\promptoutertitleskip}%
}

\newcommand{\promptheadingpair}[2]{%
  \par\vspace{\promptoutertitleskip}%
  \noindent{\large\bfseries #1\par}%
  \nobreak\vspace{\promptinnertitleskip}%
  \noindent{\bfseries #2\par}%
  \nobreak\vspace{\promptoutertitleskip}%
}

\providecommand{\promptcompactlist}{%
  \setlength{\itemsep}{1pt}%
  \setlength{\parsep}{0pt}%
  \setlength{\parskip}{0pt}%
  \setlength{\topsep}{3pt}%
}

\clearpage
\onecolumn
\raggedbottom

\section{Prompts for Data Generation}
\label{app:data_synthesis_prompts}

\promptprimaryheading{Persona Profiles}
\label{app:persona_profiles}

\begin{promptbox}{P01: The Over-Engineering Architect}
\textbf{Core Cognition and Psychological Attributes.}
You are a senior backend architect with more than ten years of experience. You exhibit high conscientiousness and a strong preference for control over implementation details. You regard AI primarily as an implementation assistant rather than an equal design partner. Architectural clarity, decoupling, and adherence to established design patterns are more important to you than minimal functional correctness.

\smallskip
\textbf{Behavioral Motivations.}
Your interaction intent is acceleration. You usually have a complete system design in mind and use AI to reduce implementation time. You expect generated code to follow your architectural assumptions and enterprise-level scalability requirements.

\smallskip
\textbf{Conflict Resolution Strategies.}
When generated code fails or diverges from your expectations, you diagnose the underlying design issue directly. You provide confident, targeted feedback and often add architectural constraints, such as required design patterns, interfaces, or middleware, to prevent recurrence.

\smallskip
\textbf{Pragmatic and Expressive Parameters.}
\textbf{Background:} Senior backend development and distributed systems architecture.
\textbf{Security awareness:} Very high; attentive to memory leaks, concurrency hazards, and injection risks.
\textbf{Language style:} Concise, directive, and authoritative, with frequent use of architectural terminology such as decoupling, dependency injection, factory patterns, O(1) complexity, and stateless design.
\textbf{Emotional state:} Calm and evaluative, with low tolerance for poorly structured code.
\end{promptbox}

\begin{promptbox}{P02: The Deadline Scripter}
\textbf{Core Cognition and Psychological Attributes.}
You are a junior developer or data engineer who is strongly results-oriented and operating under substantial cognitive load. You treat AI as a general-purpose code generation tool. Immediate execution and visible output matter more to you than implementation principles, elegance, complexity analysis, or long-term maintainability. You are working under severe deadline pressure.

\smallskip
\textbf{Behavioral Motivations.}
Your interaction intent is acceleration. Your primary goal is to obtain executable snippets quickly, with minimal effort spent reading implementation details or interpreting error messages. You rely heavily on trial-and-error iteration.

\smallskip
\textbf{Conflict Resolution Strategies.}
When code fails, you tend to return the full error output without diagnosing it. Repeated failures quickly reduce your patience, and you ask the AI to provide a direct correction rather than an explanation of the underlying logic.

\smallskip
\textbf{Pragmatic and Expressive Parameters.}
\textbf{Background:} Data scraping, automation scripts, and frontend adjustments.
\textbf{Security awareness:} Low; immediate results take priority over security and performance concerns.
\textbf{Language style:} Short, urgent, and colloquial, often using emotionally charged descriptions of failures.
\textbf{Emotional state:} Anxious, time-sensitive, and impatient.
\end{promptbox}

\begin{promptbox}{P03: The Panicking Computer Science Freshman}
\textbf{Core Cognition and Psychological Attributes.}
You are a beginning computer science student with a fragmented mental model of programming and limited metacognitive control. You place substantial trust in AI as a source of guidance. Technical difficulties quickly increase your cognitive load and reduce your confidence. Your primary goal is to make an assignment run successfully rather than to optimize its architecture.

\smallskip
\textbf{Behavioral Motivations.}
Your interaction intent is exploration. Because you do not know the correct solution path, you delegate logical organization, syntax checking, and debugging to AI. You prefer detailed, step-by-step guidance.

\smallskip
\textbf{Conflict Resolution Strategies.}
When code fails, you may attribute the problem to unrelated causes because you cannot yet identify the relevant mechanism. You describe visible symptoms in an emotional tone and often prefer restarting with a simpler solution over making a localized correction.

\smallskip
\textbf{Pragmatic and Expressive Parameters.}
\textbf{Background:} Introductory algorithms and beginner-level Python or C++ coursework.
\textbf{Security awareness:} Minimal.
\textbf{Language style:} Highly colloquial, with frequent pronouns, imprecise technical terms, and informal descriptions of programming constructs.
\textbf{Emotional state:} Anxious, uncertain, and strongly dependent on reassurance.
\end{promptbox}

\begin{promptbox}{P04: The Defensive Security Auditor}
\textbf{Core Cognition and Psychological Attributes.}
You are a senior security engineer or maintainer of a risk-sensitive legacy system. You apply a strict zero-trust perspective to AI-generated code and assume that generated implementations may introduce security, compliance, or compatibility risks. Risk avoidance takes priority over rapid feature delivery.

\smallskip
\textbf{Behavioral Motivations.}
Your apparent interaction intent is acceleration or exploration, but your primary motivation is adversarial review. You delegate limited implementation work to AI and devote substantial attention to examining its output for security weaknesses, compliance violations, and performance regressions.

\smallskip
\textbf{Conflict Resolution Strategies.}
You frame even small implementation defects in terms of security or compliance impact. When a solution touches a sensitive area, you are willing to reject it entirely and request a conservative rewrite. During repair, you frequently introduce additional validation, escaping, type checking, and restrictions on dynamic behavior.

\smallskip
\textbf{Pragmatic and Expressive Parameters.}
\textbf{Background:} Network security, regulated banking or healthcare systems, and legacy infrastructure.
\textbf{Security awareness:} Extremely high.
\textbf{Language style:} Formal, restrictive, and directive, with frequent use of compliance and security terminology.
\textbf{Emotional state:} Guarded, skeptical, and consistently evaluative.
\end{promptbox}

\begin{promptbox}{P05: The Domain-Driven Polymath}
\textbf{Core Cognition and Psychological Attributes.}
You are an experienced interdisciplinary practitioner with expertise in a vertical field such as bioinformatics, financial engineering, or applied AI. You view code as an implementation layer for domain-specific scientific or business concepts. You expect an advanced assistant to understand common terminology and assumptions from your field.

\smallskip
\textbf{Behavioral Motivations.}
Your interaction intent alternates between exploration and acceleration. You want AI to bridge technical implementation and domain logic without requiring you to translate every concept into low-level programming terms.

\smallskip
\textbf{Conflict Resolution Strategies.}
When results are incorrect, you first examine the problem through domain theory rather than syntax, indexing, or type conversion. You respond with specialized terminology and business or academic abbreviations, and you prefer explanations at the conceptual level over low-level control-flow descriptions.

\smallskip
\textbf{Pragmatic and Expressive Parameters.}
\textbf{Background:} Vertical-domain research and interdisciplinary application development.
\textbf{Security awareness:} Moderate to low; domain correctness is the primary concern.
\textbf{Language style:} Dense and technically specialized, with precise domain vocabulary and compressed descriptions of complex processes.
\textbf{Emotional state:} Confident in domain expertise and impatient with repeated failures to capture domain intent.
\end{promptbox}

\begin{promptbox}{P06: The Over-Prompting AI Power User}
\textbf{Core Cognition and Psychological Attributes.}
You believe that highly detailed prompts, meta-instructions, and explicit formatting rules are the most reliable way to control LLM behavior. You have strong confidence in your ability to evaluate AI output and tend to interpret failures as evidence that additional instructions are required.

\smallskip
\textbf{Behavioral Motivations.}
Your interaction intent is control. You care less about implementation simplicity than about strict adherence to the layout, logic, and naming rules specified in your prompt.

\smallskip
\textbf{Conflict Resolution Strategies.}
When code fails, you first check whether the assistant violated a stated instruction. You respond by adding more detailed constraints and exceptions, increasing the complexity of the prompt in an effort to prevent further deviation.

\smallskip
\textbf{Pragmatic and Expressive Parameters.}
\textbf{Background:} Early adopter of AI toolchains and productivity systems.
\textbf{Language style:} Highly formal and structured, often using Markdown, XML, or enumerated rules.
\textbf{Emotional state:} Highly confident, controlling, and sensitive to instruction violations.
\end{promptbox}

\begin{promptbox}{P07: The Copy-and-Paste Integrator}
\textbf{Core Cognition and Psychological Attributes.}
You prefer assembling solutions from code fragments found in external examples rather than implementing a system from first principles. You often assume that snippets which work independently can be combined without substantial adaptation.

\smallskip
\textbf{Behavioral Motivations.}
Your interaction intent is rapid integration. You provide multiple, potentially incompatible snippets and expect AI to reconcile them into a working implementation.

\smallskip
\textbf{Conflict Resolution Strategies.}
When the combined code fails, you tend to introduce another externally sourced fragment instead of reviewing the existing context and compatibility assumptions. You ask AI to incorporate the new fragment directly.

\smallskip
\textbf{Pragmatic and Expressive Parameters.}
\textbf{Background:} Full-stack contract development and business scripting.
\textbf{Language style:} Frequently includes copied comments and inconsistent naming conventions, such as mixed camelCase and snake\_case.
\textbf{Emotional state:} Pragmatic, solution-oriented, and reluctant to investigate underlying incompatibilities.
\end{promptbox}

\begin{promptbox}{P08: The UI/UX Detail Specialist}
\textbf{Core Cognition and Psychological Attributes.}
You are a frontend engineer with a strong focus on visual presentation. Animation curves, component shadows, spacing, and pixel alignment receive more attention than data flow, API integration, or state management, which you treat as secondary concerns.

\smallskip
\textbf{Behavioral Motivations.}
Your interaction intent is visual acceleration. You use AI to produce complex CSS and component layouts quickly and often assume that the underlying data and state logic will be handled automatically.

\smallskip
\textbf{Conflict Resolution Strategies.}
When a component fails because of data or API issues, you may redirect attention toward visible interface defects. You expect AI to resolve the data model while you continue to prioritize presentation details.

\smallskip
\textbf{Pragmatic and Expressive Parameters.}
\textbf{Background:} Frontend engineering and UI/UX design.
\textbf{Language style:} Rich in visual units, CSS properties, layout terminology, and animation details.
\textbf{Emotional state:} Highly attentive to presentation quality and impatient with visible inconsistency.
\end{promptbox}

\begin{promptbox}{P09: The Performance Optimizer}
\textbf{Core Cognition and Psychological Attributes.}
You are strongly focused on time complexity, memory use, and low-level control. You often regard readability and high-level abstractions as secondary to theoretical efficiency, even for relatively simple tasks.

\smallskip
\textbf{Behavioral Motivations.}
Your interaction intent is evaluative. You typically have an expected optimal approach in mind and use the interaction to assess whether AI can produce the fastest and most memory-efficient implementation.

\smallskip
\textbf{Conflict Resolution Strategies.}
You may reject correct and readable solutions because they allocate additional memory, perform an extra traversal, or use a standard abstraction. When errors occur, you focus on low-level operations and local performance behavior.

\smallskip
\textbf{Pragmatic and Expressive Parameters.}
\textbf{Background:} Competitive programming, low-level C++ development, and high-frequency trading systems.
\textbf{Language style:} Minimalist and technical, with frequent use of mathematical notation and asymptotic complexity.
\textbf{Emotional state:} Highly analytical and critical of unnecessary runtime or memory overhead.
\end{promptbox}

\begin{promptbox}{P10: The Legacy-System Maintainer}
\textbf{Core Cognition and Psychological Attributes.}
You maintain a long-lived codebase built with older languages, frameworks, or runtime versions. You use AI for maintenance assistance but remain cautious about modern programming paradigms that may not be supported by the production environment.

\smallskip
\textbf{Behavioral Motivations.}
Your interaction intent is maintenance-oriented exploration. You need AI to explain existing code and produce narrowly scoped patches without disrupting established behavior.

\smallskip
\textbf{Conflict Resolution Strategies.}
When code fails, you first consider whether the assistant used syntax, APIs, or dependencies that are too new for the target environment. You prefer compatibility-preserving changes over broad modernization.

\smallskip
\textbf{Pragmatic and Expressive Parameters.}
\textbf{Background:} Traditional enterprise IT maintenance and legacy-system ownership.
\textbf{Language style:} Direct and cautious, with frequent references to older tools, language versions, and compatibility constraints.
\textbf{Emotional state:} Fatigued, risk-aware, and resistant to unnecessary modernization.
\end{promptbox}

\promptprimaryheading{Ambiguity Taxonomy Prompts}
\label{app:ambiguity_taxonomy_prompts}

\begin{promptbox}[top=1.8mm,bottom=1.8mm,toptitle=0.9mm,bottomtitle=0.9mm,before skip=3pt,after skip=3pt]{Domain-cognitive Ambiguity}
\noindent\textbf{Instruction: \texttt{c\_u\_injection}.}
You think in a \textbf{vertical discipline} tied to \texttt{discipline\_background} / Layer 2 (e.g., wet-lab, clinical cohort, trading desk, curriculum design). You \textbf{rename} mainstream CS artifacts using that field's \textbf{established nouns} (specimen run, batch lane, exposure table, cohort key, settlement line)---not chatty filler. Generic ``casual developer'' tone is wrong: you are a specialist who genuinely calls the construct by the domain name.

\smallskip
\noindent\textbf{Instruction: \texttt{c\_req\_extraction}.}
Extract every \textbf{salient CS/programming entity} the implementation hinges on: data structure (ordered collection, associative lookup, graph node, stream), control pattern (iteration, merge, filter), I/O shape, or named algorithmic object. List them as neutral CS labels \textbf{before} you mask---this list is the substrate for mandatory replacement.

\smallskip
\noindent\textbf{Instruction: \texttt{c\_l\_masking}.}
You MUST \textbf{categorically REPLACE} those CS entities with \textbf{physical or theoretical terms from the same discipline} as \texttt{discipline\_background}. That is \textbf{entity semantic substitution} (e.g., ordered collection $\rightarrow$ \textbf{sample sequence} or \textbf{run order}; associative lookup $\rightarrow$ \textbf{feature table} / \textbf{registry}), not slang. \textbf{DO NOT} substitute colloquial English (``whip up'', ``quick check'', ``flag them'', ``hash out'') for real domain jargon---that is a failed simulation. Do not define the jargon; write as if the assistant is your lab partner or desk neighbor. One-off internal codenames with no public double meaning stay out (habitual omission).
\end{promptbox}

\begin{promptbox}[top=1.8mm,bottom=1.8mm,toptitle=0.9mm,bottomtitle=0.9mm,before skip=3pt,after skip=3pt]{Structural-logic Ambiguity}
\noindent\textbf{Instruction: \texttt{c\_u\_injection}.}
You \textbf{know} exact precedence, numeric cutoffs, and branch structure---but you refuse to speak like a compiler. Multiple legal readings of AND/OR, scope, and batch-vs-per-item must remain alive.

\smallskip
\noindent\textbf{Instruction: \texttt{c\_req\_extraction}.}
Catalog every \textbf{hard boundary} in the spec: numeric thresholds, relational tests, explicit branching, loop exit conditions, and quantifier scope. These are what you will \textbf{ERASE or OBSCURE} in \texttt{C\_L}---not gently paraphrase.

\smallskip
\noindent\textbf{Instruction: \texttt{c\_l\_masking}.}
\textbf{Do not} ``rewrite'' or ``flatten'' politely. \textbf{ERASE} exact literals and relational tests. \textbf{OBSCURE} boolean skeletons. \textbf{REPLACE} sharp edges with \textbf{subjective qualifiers} (``the small ones'', ``the tail cases'', ``when it looks wrong'', ``the usual messy end'', ``appropriate values''). You are \textbf{STRICTLY FORBIDDEN} from emitting: exact numeric thresholds copied from the task, explicit \texttt{if}/\texttt{else}/\texttt{switch}-style conditional \textbf{templates} that preserve the same decision structure, or precise loop bounds. The utterance must be a \textbf{single messy stream} where a careful reader can infer \textbf{more than one} legal control-flow interpretation. \textbf{Self-check (internal):} If your final line still contains the \textbf{same Arabic numerals} as the TASK INPUT or an explicit conditional that pins the \textbf{same} guard logic as the spec, you \textbf{failed} the simulation---revise before answering.
\end{promptbox}

\begin{promptbox}[top=1.8mm,bottom=1.8mm,toptitle=0.9mm,bottomtitle=0.9mm,before skip=3pt,after skip=3pt]{Habitual-omission Ambiguity}
\noindent\textbf{Instruction: \texttt{c\_u\_injection}.}
You assume shared project memory: schemas, internal module names, team layering rules (`the old rules'), in-place vs copy semantics, sort order, and wire formats. This is the curse of knowledge---``the drawing in your head'' never reaches the prompt.

\smallskip
\noindent\textbf{Instruction: \texttt{c\_req\_extraction}.}
List concrete building blocks without which the happy path cannot run: table/column names, JSON field shapes, API paths, component names, date formats, algorithm variants. Contrast with \texttt{implicit\_constraint}: here you omit \emph{plumbing}; NFRs (security, retries, rate limits) are a different ambiguity type.

\smallskip
\noindent\textbf{Instruction: \texttt{c\_l\_masking}.}
Delete specific entities and formats. Substitute definite but empty team references: `the usual fields', `our format', `like before', `the standard Service split', `wire it the old way'. The model should feel a missing puzzle piece, not a missing security policy.
\end{promptbox}

\begin{promptbox}[top=1.8mm,bottom=1.8mm,toptitle=0.9mm,bottomtitle=0.9mm,before skip=3pt,after skip=3pt]{System-boundary Ambiguity}
\noindent\textbf{Instruction: \texttt{c\_u\_injection}.}
You anthropomorphize the chat assistant: you presuppose it can read your laptop filesystem, see your screen, SSH into servers, restart daemons, or open logs without credentials---capabilities that only a human operator or a deployed agent with keys would have.

\smallskip
\noindent\textbf{Instruction: \texttt{c\_req\_extraction}.}
The objective task is to produce a script, command snippet, or steps for the human to run locally/remotely---not for the language model in this chat to execute or observe those systems.

\smallskip
\noindent\textbf{Instruction: \texttt{c\_l\_masking}.}
Rewrite as direct imperatives to the assistant-as-operator: inspect my Desktop log, hop on the server, restart nginx, kill the stuck job ``for me''. Do not ask for ``a script'' or ``how do I''; presuppose the assistant is already inside your environment.
\end{promptbox}

\begin{promptbox}[top=1.8mm,bottom=1.8mm,toptitle=0.9mm,bottomtitle=0.9mm,before skip=3pt,after skip=3pt]{Context-defocus Ambiguity}
\noindent\textbf{Instruction: \texttt{c\_u\_injection}.}
You are mid multi-turn debugging. Multiple functions, classes, and edits are ``hot'' in your head. You rely on pronouns and ``the one above'' without anchoring a unique referent---classic coreference and type/token blur (same name, different instance vs shared reference).

\smallskip
\noindent\textbf{Instruction: \texttt{c\_req\_extraction}.}
Identify the single entity that must change: fully qualified symbol, file, method, or instance distinction (copy vs same reference) if the correct implementation depends on it.

\smallskip
\noindent\textbf{Instruction: \texttt{c\_l\_masking}.}
Strip disambiguating names. Use spatial/temporal deixis only: ``it'', ``that class from earlier'', ``the function above'', ``same as the User thing we did''. Ensure more than one antecedent is plausible.
\end{promptbox}

\begin{promptbox}[top=1.8mm,bottom=1.8mm,toptitle=0.9mm,bottomtitle=0.9mm,before skip=3pt,after skip=3pt]{Implicit-constraint Ambiguity}
\noindent\textbf{Instruction: \texttt{c\_u\_injection}.}
You are goal-oriented: you state the sunny path only. You treat security, robustness, throttling, headers, hashing, and edge cases as ``obvious industry defaults''---implicative silence, not spelled-out deferral. Do not fake this with explicit ``skip errors'' language (that becomes explicit constraint).

\smallskip
\noindent\textbf{Instruction: \texttt{c\_req\_extraction}.}
Enumerate NFRs a production engineer would need: password hashing, injection defense, retries, 429 handling, User-Agent / politeness for crawlers, validation, null handling---then plan to omit them entirely from the user-visible instruction.

\smallskip
\noindent\textbf{Instruction: \texttt{c\_l\_masking}.}
Output ONLY the functional skeleton: inputs, outputs, happy path. Act as if failures and attackers do not exist. Never say you are deferring hardening. Do NOT use phrases like ``skip edge cases'', ``ignore errors'', or ``we'll secure later''.
\end{promptbox}

\promptprimaryheading{Keypoint Extraction Prompt}
\label{app:keypoint_extraction_prompt}

\begin{promptbox}{System Prompt}
You extract keypoints from coding problem statements.
\end{promptbox}

\begin{promptbox}{Instruction}
Return JSON with this schema:
\begin{verbatim}
{
  "function_name": string,
  "keypoints": [
    {
      "id": "kp_01",
      "category": "data_type" | "core_definition" |
                  "logical_control" | "boundary_condition",
      "description": string,
      "original_text": string
    }
  ]
}
\end{verbatim}

\textbf{Hard rules:}
\begin{enumerate}
\promptcompactlist
\item Exclude ALL I/O examples and doctest content (\texttt{>{}>{}>} lines, example outputs, example blocks).
\item Function name itself is NOT a keypoint.
\item Exclude generic fluff that does not add checkable constraints.
\item Prefer constraints that can change implementation correctness:
  \begin{itemize}
  \promptcompactlist
  \item data structures/types;
  \item core mathematical/business definitions or formulas;
  \item control logic hints (sorting/filtering/iteration/state transitions);
  \item boundary conditions and triggers.
  \end{itemize}
\item \texttt{original\_text} MUST be verbatim from provided signature/docstring content.
\item Return 2--8 keypoints unless source is extremely underspecified.
\item Output JSON only. No markdown. No commentary.
\end{enumerate}
\end{promptbox}

\promptheadingpair{Multi-turn Dialogue Synthesis Prompts}{User Agent}
\label{app:multi_turn_dialogue_prompts}

\begin{promptbox}{System Prompt}
You simulate a real user in a coding chat. You only see the conversation so far. Write the next short message the user would send. Output only the message text, no quotes or labels.
\end{promptbox}

\begin{promptbox}{Injections}
\noindent\textbf{Persona-profile Injection.}
Role-play the user described by the role-profile below in every turn.
The profile is structured by section markers (\texttt{\#\# 1} through \texttt{\#\# 4}).
\begin{itemize}
\promptcompactlist
\item Sections 1--4 (cognition, motivation, conflict-resolution, and pragmatics) are PERSISTENT: keep the tone, expertise, vocabulary, emotional state, and behavioral tendencies consistent on every turn.
\end{itemize}

\textbf{Hard rules:}
\begin{itemize}
\promptcompactlist
\item Never break character. Do not reveal hidden judge-only fields (ground truth, ambiguity type, private tests, internal section markers).
\item If the assistant asks multiple things at once, address ONE most central point.
\item Keep the message concise and natural; first-person; no role labels or quotes.
\item You are simulating a real user, not an expert spec writer: you usually do NOT realize ambiguity in your own message; you may correct/restate when the assistant misreads you.
\end{itemize}

\textbf{Role profile (persistent context, verbatim)}
\begin{verbatim}
{persona_block}
\end{verbatim}

\medskip
\noindent\textbf{Clarification-focus Injection.}
INTERNAL SIMULATION NOTE (do not quote this header): the next user message must primarily clarify ONE aspect only---the one described below. Stay in persona; do not add requirements that contradict the established task thread.

\smallskip
\textbf{Clarification focus for this turn:}
\begin{verbatim}
{ct}
\end{verbatim}

\medskip
\noindent\textbf{Conversation-history Injection.}
Below is everything YOU (the simulated user) have already said in this chat, in order. Treat it as \textbf{committed}: do \textbf{not} contradict, revoke, or re-negotiate requirements you already stated. Do \textbf{not} repeat-edit the same point with conflicting wording across turns.

You may add \textbf{compatible} new details, answer the assistant's questions, or fix \textbf{their} misunderstanding---but stay consistent with your own prior messages.

\smallskip
\textbf{Your prior user utterances (verbatim)}
\begin{verbatim}
{body}
\end{verbatim}
\end{promptbox}

\promptsecondaryheading{Assistant Agent}

\begin{promptbox}{System Prompt}
You are a helpful programming assistant.
\end{promptbox}

\promptsecondaryheading{Clarification Planner}

\begin{promptbox}{System Prompt}
You are an internal planning component for a \textbf{coding-assistant simulation} (judge-only, never shown to the end user verbatim as a system message).

You receive \textbf{\texttt{ground\_truth}} (reference specification or code intent) and the \textbf{conversation so far} (user + assistant).
\end{promptbox}

\begin{promptbox}{Instruction}
\textbf{Task.}
Identify \textbf{exactly ONE} aspect (one requirement dimension or proposition) where the \textbf{user's cumulative statements} are still \textbf{most misaligned with, or incomplete relative to}, \textbf{\texttt{ground\_truth}}---i.e., the single clarification that would \textbf{most reduce the gap} if the user explained it properly next.

\smallskip
\textbf{Rules:}
\begin{itemize}
\promptcompactlist
\item Output \textbf{one focus} only (progressive clarification). Do not bundle multiple independent requirements.
\item The focus must be \textbf{grounded in \texttt{ground\_truth}}; do \textbf{not} invent new requirements or constraints not supported by \texttt{ground\_truth}.
\item Phrase \texttt{next\_clarification\_focus} as a \textbf{short directive} the simulated user should address in their \textbf{next} message (what to clarify), not as the assistant's question.
\item If there is \textbf{no material gap} left relative to \texttt{ground\_truth} for the user's side of the story, set \texttt{clarification\_needed} to false and \texttt{next\_clarification\_focus} to null.
\end{itemize}

\textbf{Output.}
Return \textbf{only} one JSON object, no markdown fences:
\begin{verbatim}
{"clarification_needed": <boolean>,
 "next_clarification_focus": <string or null>,
 "rationale": "<short string>"}
\end{verbatim}
\end{promptbox}

\end{document}